\documentclass[10pt,twocolumn,letterpaper]{article}

\usepackage{cvpr}    
\usepackage{multirow}
\usepackage{graphicx}
\usepackage{booktabs}

\usepackage[dvipsnames]{xcolor}

\definecolor{cvprblue}{rgb}{0.21,0.49,0.74}
\usepackage[pagebackref,breaklinks,colorlinks,citecolor=cvprblue]{hyperref}

\def\paperID{67} 
\def\confName{CVPR}
\def\confYear{2024}

\title{Achieving Reliable and Fair Skin Lesion Diagnosis via Unsupervised Domain Adaptation}

\author{Janet Wang, Yunbei Zhang, Zhengming Ding, Jihun Hamm \\
Tulane University\\
{\tt\small \{swang47, yzhang111, zding1, jhamm3\}@tulane.edu}
}

\begin{document}
\maketitle
\begin{abstract}
The development of reliable and fair diagnostic systems is often constrained by the scarcity of labeled data. To address this challenge, our work explores the feasibility of unsupervised domain adaptation (UDA) to integrate large external datasets for developing reliable classifiers. The adoption of UDA with multiple sources can simultaneously enrich the training set and bridge the domain gap between different skin lesion datasets, which vary due to distinct acquisition protocols. Particularly, UDA shows practical promise for improving diagnostic reliability when training with a custom skin lesion dataset, where only limited labeled data are available from the target domain. In this study, we investigate three UDA training schemes based on source data utilization: single-source, combined-source, and multi-source UDA. Our findings demonstrate the effectiveness of applying UDA on multiple sources for binary and multi-class classification. A strong correlation between test error and label shift in multi-class tasks has been observed in the experiment. Crucially, our study shows that UDA can effectively mitigate bias against minority groups and enhance fairness in diagnostic systems, while maintaining superior classification performance. This is achieved even without directly implementing fairness-focused techniques. This success is potentially attributed to the increased and well-adapted demographic information obtained from multiple sources.
\end{abstract}    
\begin{figure*}[tbp]
    \centering
    \includegraphics[width=13cm]{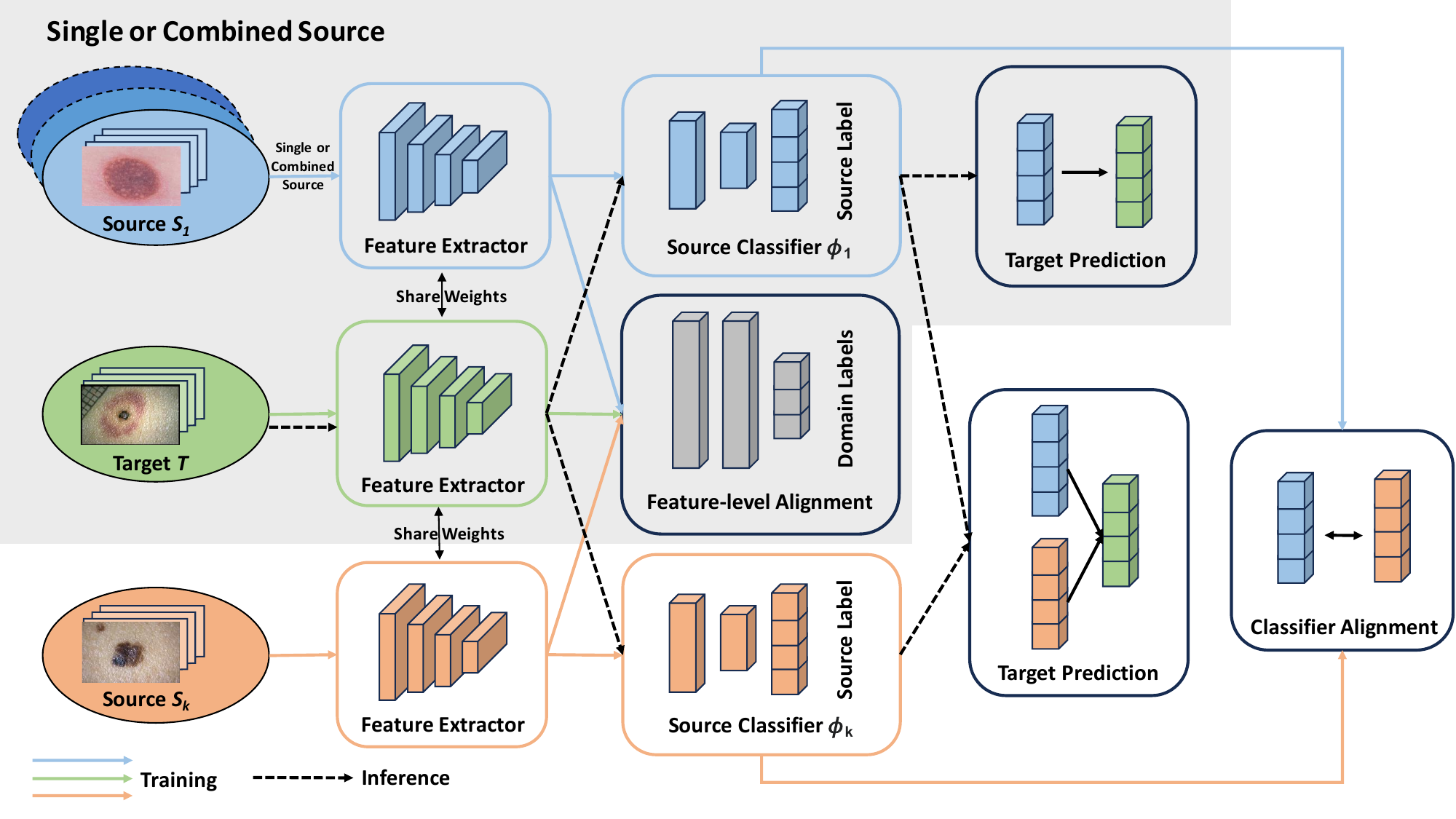}
    \caption{Illustration of single-, combined-, and multi-source UDA training for skin lesion classification. The shaded area demonstrates how single- and combined-source UDA operate. Specifically, combined-source UDA involves an additional step of aggregating multiple datasets into a single source, after which it follows the same training procedure as single-source UDA. The entire figure illustrates the multi-source training scheme, where all source domains (marked in orange and blue) are aligned with the target domain (marked in green) via a domain alignment component (marked in grey). A classifier is trained on each source domain independently, while also aligning with those trained on other source domains. All classifiers will be used to make target predictions during inference.
    }
    \label{fig:uda}
\end{figure*}

\section{Introduction}
AI-assisted diagnostic systems have shown expert-level proficiency in classifying skin cancers, conditions typically identified through visual diagnoses \cite{Esteva2017DermatologistlevelCO, Liu2020-vx, Brinkerarticle}. Such systems can potentially contribute to teledermatology as diagnostic and decision-support tools, enhancing dermatological access in rural areas where medical resources are scarce \cite{Coustasse2019UseOT}. However, the development of reliable diagnostic models is often challenged by the scarcity of labeled images—a common issue in medical imaging analysis. Efforts to address data scarcity, such as leveraging \textit{Generative Adversarial Networks} (GANs) to augment training sets \cite{pmlr-v116-ghorbani20a, bissoto2021gan, ren2021improve}, also demand a large volume of labeled data to achieve ideal performance.

Our motivation stems from a prevalent problem in the medical imaging analysis: the development of classifiers for custom medical datasets often faces a lack of sufficient labeled data, with the target domain distribution remaining unknown. To navigate this challenge, public labeled skin lesion datasets can be utilized to enrich the training set's information and diversity. Each dataset represents a distinct domain, due to varying image acquisition protocols and sources. Therefore, effectively managing the domain shift between source and target domains is essential to ensure that features learned from source domains can be successfully applied to target domains, with limited target labeled data available as reference \cite{Yosinski2014HowTA, Ovadia2019CanYT}.

Unsupervised domain adaptation (UDA) allows deep neural networks to learn features discriminative between classes and invariant across domains by jointly optimizing underlying features and two components: the label predictor and the domain alignment component \cite{pmlr-v37-ganin15}. Relevant studies \cite{CHAMARTHI2024101430, fogelberg2023domain} have validated UDA's efficacy for binary melanoma-nevus classification on dermoscopic images. Notably, \cite{CHAMARTHI2024101430} focuses on the investigation of various UDA methods in a single-source, single-target, homogeneous adaptation context, revealing the prominent effectiveness of adversarial-based UDA in binary classification.

Unlike previous research, our study seeks to explore the potential of UDA in maximizing the utility of diverse, public skin lesion data sources through different training schemes: single-source, combined-source, and multi-source UDA. Following \cite{CHAMARTHI2024101430}, we employ two representative adversarial-based UDA methods, ADDA and DANN, to investigate single-source and combined-source training. Given the absence of prior research on multi-source UDA in skin lesion classification, we have chosen two methods representing moment-matching and adversarial-based UDA approaches for multi-source training, M$^3$SDA and MSDN, respectively.

In our experiments, we leverage multiple skin lesion datasets that comprise dermoscopic or clinical images. For each target dataset, we either use a single source or leverage all sources to train a classifier with UDA, under the challenging assumption that labeled data from target domain is inaccessible. If a single-source domain is considered, a corresponding UDA method is applied to adapt it to the target domain. Alternatively, if multiple source domains are considered, we may either combine them for single-source UDA or employ multi-source UDA methods designed for aligning multiple sources. We explore all training schemes in binary and multi-class classification tasks, across the six target sets. Given the richer and well-aligned demographic feature achievable through UDA with multiple training sources, we anticipate a reduced performance gap between groups differentiated by demographic attributes, such as age and skin type. Consequently, we assess the impact of these UDA training schemes on fairness using various metrics. We believe our work holds significant practical value, offering a promising approach to building a reliable diagnostic system on custom datasets, which often suffer from limitation in size and diversity. The key contributions of our work are three-fold:\\
\indent 1. To the best of our knowledge, this is the first systematic evaluation of single- and multi-source UDA methods applied to skin lesion classification across six datasets, for both binary and multi-class tasks.\\
\indent 2. We provide a detailed, quantitative analysis of UDA's performance in skin lesion classification and find that applying UDA with multiple sources consistently outperforms its single-source counterpart across target domains, for both binary and multi-class classification.\\
\indent 3. To the best of our knowledge, we are the first to assess the effect of UDA on fairness in skin lesion classification. Our analysis reveals that leveraging multiple sources with UDA can effectively ameliorate bias against underrepresented groups.

\section{Related Works}\label{sec:background}
\subsection{Unsupervised Domain Adaptation}
UDA methods aim to mitigate the domain shifts between source and related, yet unlabeled target domains. Existing UDA methods can be categorized based on their knowledge transfer strategies into two primary groups: image alignment and feature alignment. Image alignment uses GANs to translate source images into those that are visually similar to the target domain. In contrast, feature alignment approaches distribution shifts at the feature level. Domain Adversarial Neural Network (DANN) and Adversarial Discriminative Domain Adaptation (ADDA) exemplify adversarial training methods that learn invariant features across domains \cite{ganin2015unsupervised, tzeng2017adversarial}. There's a growing interest in adapting multiple resources to an unlabeled domain. For instance, Multi-source Domain Adversarial Networks (MDAN) enhance domain adaptation by optimizing task-adaptive generalization bounds \cite{zhao2018adversarial}. Moreover, the Moment Matching for Multi-Source Domain Adaptation (M$^3$SDA) method transfers knowledge learned from multiple labeled sources to an unlabeled target domain by dynamically aligning moments of their feature distributions \cite{peng2019moment}.

\subsection{Domain Adaptation in Skin Lesion Analysis}
Prior works have validated the feasibility of DA to enhance performance in skin lesion classification. \cite{gu2019progressive}, for example, adopts CycleGAN as a domain adaptation technique to perform invariant attribute translation between skin disease datasets at image level. In their work, source and target domains are aligned to include the same disease classes, under the assumption that target labels are known. On the other hand, \cite{fogelberg2023domain} focuses on applying single-source UDA to dermoscopic images for melanoma-nevus classification, relaxing the assumption about the target domain. Extending this work, \cite{CHAMARTHI2024101430} investigates various single-source UDA and finds that adversarial-based UDA training is effective for melanoma-nevus classification.

\subsection{Fairness in Skin Lesion Analysis}
Recent studies have revealed bias in AI-based diagnosis systems against underrepresented sub-populations \cite{Larrazabal2020GenderII, seyyed2021underdiagnosis}. This bias also manifests in skin lesion analysis. Studies have shown that training data with limited skin tone diversity can lead to under-diagnosis of skin cancers in people with darker skin tones, potentially delaying critical treatment and hindering the reliable deployment of AI in teledermatology \cite{daneshjou2022disparities, groh2021evaluating}. Furthermore, research suggests that these systems can exhibit performance gaps across different age and gender subgroups, raising concerns about under- and over-diagnosis across groups \cite{fogelberg2023domain}. While collecting and annotating more demographic-diverse data is ideal, it's not always practical due to resource limitations. Nevertheless, addressing fairness in AI-based diagnosis still remains crucial in today’s skin lesion analysis.

\section{Datasets and Methods}\label{sec:methods}
\subsection{Dataset}
We evaluate the three UDA training schemes using six public and well-annotated skin lesion datasets, consisting of either dermoscopic or clinical images, as shown in Table \ref{dataset_description_table}. We specifically separate the dermoscopic and clinical pairs in the Derm7pt dataset into two independent datasets—Derm7pt-derm and Derm7pt-clinic—due to their differing image types. Example images for each dataset can be found in Appendix \cref{sec:demo_imgs}. We focus on the most common eight skin cancer conditions, whose diagnosis is critical for early screening. Class distributions are detailed in Table \ref{class_distribution_table}. 

We adhere to the official data split for Derm7pt, and adopt a 0.2 test split ratio with class stratification for the remaining datasets. For the Fitzpatrick17k subset considered, we remove duplicate or non-lesion data and apply region-of-interest (ROI) preprocessing to the images, using ROI-cropped images in subsequent experiments. Details about the ROI-cropped images can be found in Appendix \cref{sec:roi}. Since our objective is to assess the feasibility of UDA in skin lesion analysis, we resize each image to a $64 \times 64$ pixel format to facilitate the experiments. Throughout the remainder of the paper, we will refer to these datasets by their abbreviated names as listed in Table \ref{dataset_description_table}.

\begin{table}[!tbp]
\centering
\resizebox{0.5\textwidth}{!}{%
\begin{tabular}{@{}llllp{5.5cm}@{}}
\toprule
Dataset & Abbrev. & Samples & Type & Diagnostic Classes \\
\midrule

ISIC 2018 \cite{codella2018skin, tschandl2018ham10000} & isic2018 & 9688 & Dermoscopic & Melanocytic nevus, melanoma, basal cell carcinoma, benign keratosis, dermatofibroma, vascular lesion \\

ISIC 2020 \cite{Rotemberg2021-az} & isic2020 & 6000 & Dermoscopic & Melanocytic nevus, melanoma, benign keratosis \\

PAD-UFES-20 \cite{Pacheco2020-nl} & pad & 2298 & Clinical & Melanocytic nevus, melanoma, basal cell carcinoma, benign keratosis, actinic keratosis, squamous cell carcinoma \\

Fitzpatrick17k \cite{groh2021evaluating} & fitz-roi & 1496 & Clinical & Melanocytic nevus, melanoma, basal cell carcinoma, benign keratosis, actinic keratosis, squamous cell carcinoma, dermatofibroma \\

Derm7pt-Derm \cite{kawahara2018seven} & d7pt-d & 803 & Dermoscopic & Melanocytic nevus, melanoma, basal cell carcinoma, benign keratosis, dermatofibroma, vascular lesion \\

Derm7pt-Clinic \cite{kawahara2018seven} & d7pt-c & 803 & Clinical & Melanocytic nevus, melanoma, basal cell carcinoma, benign keratosis, dermatofibroma, vascular lesion \\
\bottomrule
\end{tabular}%
}
\caption{Datasets description.}
\label{dataset_description_table}
\end{table}

\begin{table}[!t]
\centering
\scriptsize
\resizebox{0.45\textwidth}{!}{%
\begin{tabular}{lcccccc}
\toprule
~ & isic2018 & isic2020 & pad & fitz-roi & d7pt-d/c  \\ 
\midrule

NEV & 6705 (69) & 5193 (86) & 244 (11) & 140 (9) & 475 (59)  \\ 

MEL & 1113 (12) & 584 (10) & 52 (2) & 299 (20) & 191 (24)  \\ 

BCC & 514 (5) & - & 845 (37) & 416 (28) & 35 (4)  \\ 

BKL & 1099 (11) & 223 (4) & 235(10) & 52(4) & 67(8)  \\ 

AK & - & - & 730 (32) & 104 (7) & - \\ 

SCC & - & - & 192 (8) & 408 (27) & -  \\  

DF & 115 (1) & - & - & 77 (5) & 13 (2)  \\  

VL & 142 (2) & - & - & - & 22 (3) \\  
\bottomrule
\end{tabular}%
}
\caption{Class distribution for each dataset. The numbers enclosed in parentheses represent percentages.}
\label{class_distribution_table}
\end{table}

\begin{table*}[tbp]
    \centering
    \scriptsize
    \resizebox{0.9\textwidth}{!}{%
    \begin{tabular}{l l c c c c c c |c}
    \toprule
        \multicolumn{2}{c}{Source} & isic2018 & isic2020 & pad & fitz-roi & d7pt-d & d7pt-c & average\\ 
        \midrule
        
        \multirow{6}{*}{single}
        & isic2018 & - & 86.8$\pm$\tiny{1.8} & 60.0$\pm$\tiny{1.5} & 81.5$\pm$\tiny{1.7} & 80.3$\pm$\tiny{1.6} & 70.8$\pm$\tiny{2.3} & 75.9$\pm$\tiny{2.0} \\
        & isic2020 & 86.6$\pm$\tiny{1.7} & - & 78.4$\pm$\tiny{1.6} & 67.7$\pm$\tiny{1.9} & 60.4$\pm$\tiny{2.1} & 58.5$\pm$\tiny{2.2} & 70.3$\pm$\tiny{1.8} \\
        & pad & 67.1$\pm$\tiny{2.0} & 61.2$\pm$\tiny{1.8} & - & 65.6$\pm$\tiny{1.6} & 77.5$\pm$\tiny{2.1} & 56.9$\pm$\tiny{2.3} & 65.7$\pm$\tiny{1.7} \\
        & fitz-roi & 71.1$\pm$\tiny{2.2} & 66.5$\pm$\tiny{2.0} & 58.8$\pm$\tiny{1.7} & - & 72.4$\pm$\tiny{1.8} & 70.8$\pm$\tiny{2.1} & 67.9$\pm$\tiny{2.0} \\
        & d7pt-d & 78.3$\pm$\tiny{1.9} & 72.6$\pm$\tiny{1.7} & 55.1$\pm$\tiny{2.3} & 72.7$\pm$\tiny{1.8} & - & 69.0$\pm$\tiny{2.2} & 69.5$\pm$\tiny{1.8} \\
        & d7pt-c & 71.9$\pm$\tiny{2.1} & 67.3$\pm$\tiny{1.9} & 74.1$\pm$\tiny{2.2} & 62.1$\pm$\tiny{2.0} & 78.5$\pm$\tiny{2.0} & - & 70.8$\pm$\tiny{2.1} \\ 
        \cmidrule{2-9}

        & average & 75.0$\pm$\tiny{2.0} & 70.9$\pm$\tiny{1.8} & 65.3$\pm$\tiny{1.9} & 69.9$\pm$\tiny{1.8} & 73.8$\pm$\tiny{1.9} & 65.2$\pm$\tiny{2.2} & - \\
        \midrule

        \multirow{6}{*}{single-DANN}
        &isic2018 & - & 86.8$\pm$\tiny{2.3} & 72.2$\pm$\tiny{3.4} & 85.5$\pm$\tiny{2.8} & 81.8$\pm$\tiny{3.1} & 73.1$\pm$\tiny{2.7} & 79.9$\pm$\tiny{3.0}  \\ 
        &isic2020 & 79.9$\pm$\tiny{2.6} & - & 82.6$\pm$\tiny{3.1} & 84.2$\pm$\tiny{3.3} & 74.8$\pm$\tiny{2.9} & 64.4$\pm$\tiny{2.5} & 77.2$\pm$\tiny{2.8}  \\ 
        &pad & 63.5$\pm$\tiny{3.1} & 64.2$\pm$\tiny{2.8} & - & 83.2$\pm$\tiny{2.5} & 71.5$\pm$\tiny{2.7} & 62.4$\pm$\tiny{3.3} & 69.0$\pm$\tiny{2.9}  \\ 
        &fitz-roi & 70.7$\pm$\tiny{3.0} & 68.6$\pm$\tiny{3.2} & 72.5$\pm$\tiny{3.3} & - & 78.0$\pm$\tiny{2.9} & 73.2$\pm$\tiny{3.1} & 72.6$\pm$\tiny{3.1}  \\ 
        &d7pt-d & 71.7$\pm$\tiny{2.9} & 73.0$\pm$\tiny{3.2} & 68.6$\pm$\tiny{2.8} & 84.8$\pm$\tiny{2.7} & - & 72.5$\pm$\tiny{3.0} & 74.1$\pm$\tiny{2.8}  \\ 
        &d7pt-c & 75.4$\pm$\tiny{3.2} & 64.2$\pm$\tiny{3.1} & 73.9$\pm$\tiny{2.6} & 83.7$\pm$\tiny{3.1} & 81.5$\pm$\tiny{3.3} & - & 75.7$\pm$\tiny{3.0}  \\ 
        \cmidrule{2-9}
        
        &average & 72.3$\pm$\tiny{3.1} & 71.4$\pm$\tiny{3.0} & 74.0$\pm$\tiny{3.2} & 84.2$\pm$\tiny{2.9} & 77.5$\pm$\tiny{2.8} & 69.1$\pm$\tiny{3.3} & - \\ 
        \midrule

        \multirow{6}{*}{combined}
        & ERM & 85.9$\pm$\tiny{1.7} & {87.1$\pm$\tiny{2.2}} & 74.0$\pm$\tiny{1.9} & 76.5$\pm$\tiny{2.3} & 76.7$\pm$\tiny{2.1} & 74.0$\pm$\tiny{2.4} & 79.0$\pm$\tiny{2.0} \\
        
        & rs-ERM & 86.3$\pm$\tiny{2.1} & 87.8$\pm$\tiny{2.0} & 74.9$\pm$\tiny{2.3} & 84.0$\pm$\tiny{2.1} & 80.5$\pm$\tiny{2.4} & 74.6$\pm$\tiny{2.2} & 81.4$\pm$\tiny{2.2} \\
        
        & DANN & \textbf{86.9$\pm$\tiny{2.3}} & 85.4$\pm$\tiny{1.9} & 70.0$\pm$\tiny{2.1} & 87.7$\pm$\tiny{2.2} & 82.7$\pm$\tiny{2.0} & 74.9$\pm$\tiny{2.3} & 81.3$\pm$\tiny{2.1}\\

        & rs-DANN& 86.4$\pm$\tiny{2.1} & 87.3$\pm$\tiny{2.3} & 76.9$\pm$\tiny{2.2} & \textbf{92.2$\pm$\tiny{2.3}} & \textbf{85.5$\pm$\tiny{2.2}} & 74.3$\pm$\tiny{2.4} & 83.8$\pm$\tiny{2.2}\\
        
        & ADDA & 84.4$\pm$\tiny{1.9} & 84.2$\pm$\tiny{2.0} & 73.9$\pm$\tiny{2.2} & 83.3$\pm$\tiny{1.9} & 78.0$\pm$\tiny{2.1} & 72.8$\pm$\tiny{2.3} & 79.4$\pm$\tiny{2.1} \\

        & rs-ADDA& 86.1$\pm$\tiny{2.3} & 85.2$\pm$\tiny{2.1} & 79.8$\pm$\tiny{2.4} & 79.2$\pm$\tiny{2.0} & 82.7$\pm$\tiny{2.2} & 71.2$\pm$\tiny{2.1} & 80.7$\pm$\tiny{2.2} \\
        \midrule
        
        \multirow{4}{*}{multi} 
        & MDAN & 84.3$\pm$\tiny{3.6} & 85.7$\pm$\tiny{3.5} & 70.8$\pm$\tiny{3.2} & 88.6$\pm$\tiny{3.3} & 81.8$\pm$\tiny{3.5} & 74.8$\pm$\tiny{3.1} & 81.0$\pm$\tiny{3.4}\\

        & rs-MDAN& 86.4$\pm$\tiny{3.4} & 85.6$\pm$\tiny{3.9} & 76.3$\pm$\tiny{3.6} & 84.4$\pm$\tiny{3.8} & 82.1$\pm$\tiny{3.8} & 75.1$\pm$\tiny{3.4} & 81.7$\pm$\tiny{3.6}\\
        
        & M$^3$SDA & 85.7$\pm$\tiny{3.3} & 85.7$\pm$\tiny{3.6} & \textbf{86.4$\pm$\tiny{3.5}} & 77.5$\pm$\tiny{3.4} & 74.5$\pm$\tiny{3.7} & 71.6$\pm$\tiny{3.6} & 80.2$\pm$\tiny{3.6}\\

        & rs-M$^3$SDA & 86.3$\pm$\tiny{3.5} & \textbf{89.4}$\pm$\tiny{3.8} & 84.9$\pm$\tiny{3.4} & 83.4$\pm$\tiny{3.8} & 85.1$\pm$\tiny{3.6} & \textbf{78.0$\pm$\tiny{3.7}} & 84.5$\pm$\tiny{3.7} \\


    \bottomrule
    \end{tabular}%
}
    \caption{AUROC for melanoma-nevus binary classification, with each column showing results for a target domain. Single-source training treats each non-target dataset as a source. Combined-source aggregates non-target datasets, applying single-source UDA thereafter. Multi-source results are derived without intentional dataset aggregation, instead relying on UDA methods designed to adapt multiple sources simultaneously. "rs-" denotes the inclusion of a weighted random sampler.}
    \label{binary_auroc_table}
\end{table*}

\subsection{UDA Methods}
In this study, we investigate single-source, combined-source, and multi-source UDA training strategies, which are illustrated in Fig. \ref{fig:uda}. Prior research using adversarial training-based UDA achieved promising results in classifying melanoma and nevus on single-source dermatoscopic images \cite{CHAMARTHI2024101430}. Inspired by this, we selected DANN and ADDA, two representative adversarial-based UDA methods from that study, to investigate combined-source adaptation. Since DANN and ADDN have demonstrated comparable performance in single-source adaptation, we only test single-source UDA with DANN. 

Existing literature hasn't explored the application of multi-source UDA to the area of skin lesion analysis. Therefore, we selected two methods representing moment-matching and adversarial-based UDA approaches for multi-source training, M$^3$SDA and MSDN, respectively. It's important to note that a comprehensive comparison of various UDA methods falls outside the scope of this work. Our focus here is on evaluating the effectiveness of different UDA training schemes (i.e., single-source, combined-source, and multi-source UDA) in integrating available public data resources and bridging the gap between their domains, for skin lesion classification.

\subsection{Evaluating Fairness}
Adapting the methods from \cite{du2022fairdisco, du2020fairness}, fairness is quantified using three metrics: Predictive Quality Disparity (\textbf{PQD}), Demographic Disparity Metric (\textbf{DPM}), and Equality of Opportunity Metric (\textbf{EOM}). Let $S$ be the sensitive group, for example, $S =\{\mathrm{male}, \mathrm{female}\}$ and denote the label set by $\{1, ..., M\}$, we can formulate the three metrics as follow:

\begin{enumerate}
    \item Predictive Quality Disparity (\textbf{PQD}) measures the prediction quality difference between each sensitive groups:
$$
\mathrm{PQD} = \frac{\min(\mathrm{acc}_j, j\in S)}{\max(\mathrm{acc}_j, j \in S)}
$$
here $\mathrm{acc}_j$ is the accuracy (or AUROC) of the data in $j$-th sensitive group.

    \item Demographic Disparity Metric (\textbf{DPM}) computes the percentage diversities of positive outcomes for each sensitive group.
$$
\mathrm{DPM} = \frac{1}{M} \sum_{i=1}^M \frac{\min[\;p(\hat y = i|s=j), j\in S\;]}{\max[\;p(\hat y =i | s= j), j \in S\;]}
$$
where $\hat y$ is the prediction of the model.

    \item Equality of Opportunity Metric (\textbf{EOM}) requires that different sensitive group should have similar true positive rates.
$$
\mathrm{EOM}=\frac{1}{M} \sum_{i=1}^M \frac{\min[\;p(\hat y = i|y=i, s=j), j\in S\;]}{\max[\;p(\hat y =i |y=i, s= j), j \in S\;]}
$$
where $y$ is the true label.
\end{enumerate}

\begin{table*}[!tbp]
\centering
\scriptsize
\resizebox{0.9\textwidth}{!}{%
\begin{tabular}{l l *{6}{c} |c}
 \toprule

        \multicolumn{2}{c}{Source} & isic2018 & isic2020 & pad & fitz-roi & d7pt-d & d7pt-c & average \\
        \midrule        
        
        \multirow{6}{*}{single} 
        & isic2018 & - & 86.5$\pm$\tiny{3.8} & 23.9$\pm$\tiny{3.1} & 21.3$\pm$\tiny{2.5} & 66.9$\pm$\tiny{3.9} & 54.2$\pm$\tiny{2.7} & 50.6$\pm$\tiny{3.6} \\ 
        & isic2020 & 70.4$\pm$\tiny{4.2} & - & 10.4$\pm$\tiny{3.4} & 21.0$\pm$\tiny{4.1} & 47.6$\pm$\tiny{2.8} & 46.1$\pm$\tiny{4.4} & 39.1$\pm$\tiny{3.9} \\ 
        & pad & 61.1$\pm$\tiny{2.9} & 63.7$\pm$\tiny{3.6} & - & 30.0$\pm$\tiny{2.8} & 45.3$\pm$\tiny{4.3} & 44.5$\pm$\tiny{3.2} & 48.9$\pm$\tiny{3.8} \\ 
        & fitz-roi & 56.9$\pm$\tiny{4.3} & 60.2$\pm$\tiny{3.7} & 38.3$\pm$\tiny{3.5} & - & 48.6$\pm$\tiny{3.1} & 46.6$\pm$\tiny{4.2} & 50.1$\pm$\tiny{4.4} \\ 
        & d7pt-d & 64.9$\pm$\tiny{4.0} & 69.1$\pm$\tiny{2.9} & 20.4$\pm$\tiny{4.4} & 21.0$\pm$\tiny{3.6} & - & 54.2$\pm$\tiny{2.5} & 45.9$\pm$\tiny{3.9} \\ 
        & d7pt-c & 67.9$\pm$\tiny{4.5} & 78.5$\pm$\tiny{4.1} & 13.3$\pm$\tiny{3.8} & 23.3$\pm$\tiny{4.3} & 60.8$\pm$\tiny{2.9} & - & 48.8$\pm$\tiny{3.4} \\ 
        \cmidrule{2-9}
        & average & 64.2$\pm$\tiny{3.2} & 71.6$\pm$\tiny{4.3} & 21.3$\pm$\tiny{3.9} & 23.3$\pm$\tiny{3.6} & 53.8$\pm$\tiny{4.0} & 49.1$\pm$\tiny{3.5} & 47.2$\pm$\tiny{4.1} \\ 
        \midrule
        
        \multirow{6}{*}{combined}
        & ERM & 69.7$\pm$\tiny{2.7} & 86.8$\pm$\tiny{2.8} & 35.2$\pm$\tiny{2.2} & 33.3$\pm$\tiny{2.6} & 64.4$\pm$\tiny{2.1} & 58.0$\pm$\tiny{2.3} & 57.9$\pm$\tiny{2.9} \\ 
        & rs-ERM & 72.6$\pm$\tiny{3.8} & 88.7$\pm$\tiny{4.0} & 43.3$\pm$\tiny{3.7} & 36.7$\pm$\tiny{3.4} & \textbf{69.7$\pm$\tiny{4.0}} & 57.8$\pm$\tiny{4.3} & 61.4$\pm$\tiny{3.8} \\
        & DANN & 69.5$\pm$\tiny{3.6} & 86.1$\pm$\tiny{4.0} & 33.9$\pm$\tiny{3.1} & 33.0$\pm$\tiny{3.9} & 61.3$\pm$\tiny{3.5} & 56.0$\pm$\tiny{3.2} & 56.6$\pm$\tiny{3.6} \\
        & rs-DANN & 72.8$\pm$\tiny{3.5} & \textbf{88.8$\pm$\tiny{3.9}} & \textbf{45.9$\pm$\tiny{4.3}} & 37.7$\pm$\tiny{3.8} & 64.1$\pm$\tiny{3.3} & 61.8$\pm$\tiny{3.6} & 61.8$\pm$\tiny{4.0} \\
        & ADDA & 68.7$\pm$\tiny{4.2} & 86.6$\pm$\tiny{3.9} & 35.7$\pm$\tiny{4.3} & 19.3$\pm$\tiny{3.7} & 65.7$\pm$\tiny{4.0} & 58.0$\pm$\tiny{3.9} & 55.7$\pm$\tiny{4.1} \\ 
        & rs-ADDA & \textbf{74.0$\pm$\tiny{3.6}} & 86.9$\pm$\tiny{4.3} & 40.7$\pm$\tiny{3.2} & 38.7$\pm$\tiny{3.8} & 67.9$\pm$\tiny{3.5} & \textbf{63.4$\pm$\tiny{3.9}} & 61.9$\pm$\tiny{3.6} \\
        \midrule
        \multirow{4}{*}{multi} 
        & MDAN & 66.7$\pm$\tiny{3.8} & 83.6$\pm$\tiny{4.1} & 38.7$\pm$\tiny{4.2} & 36.3$\pm$\tiny{3.9} & 60.8$\pm$\tiny{3.7} & 53.7$\pm$\tiny{4.3} & 56.6$\pm$\tiny{3.8} \\
        & rs-MDAN & 70.2$\pm$\tiny{4.0} & 86.7$\pm$\tiny{4.3} & 38.3$\pm$\tiny{3.6} & \textbf{40.7$\pm$\tiny{3.9}} & 65.7$\pm$\tiny{4.1} & 60.0$\pm$\tiny{3.8} & 60.3$\pm$\tiny{4.2} \\
        & M$^3$SDA & 67.4$\pm$\tiny{3.6} & 81.2$\pm$\tiny{4.3} & 38.0$\pm$\tiny{3.8} & 38.7$\pm$\tiny{4.0} & 63.8$\pm$\tiny{4.1} & 59.1$\pm$\tiny{3.9} & 58.0$\pm$\tiny{3.7} \\
        & rs-M$^3$SDA & 71.3$\pm$\tiny{3.9} & 86.6$\pm$\tiny{3.7} & 40.7$\pm$\tiny{3.5} & 40.0$\pm$\tiny{4.1} & 65.9$\pm$\tiny{4.3} & 62.3$\pm$\tiny{4.0} & 61.1$\pm$\tiny{3.9} \\
\bottomrule
\end{tabular}%
}
\caption{Accuracy of 8-label classification. The eight skin conditions are listed in Table \ref{class_distribution_table}.}
\label{8class_accuracy_table}
\end{table*}

\subsection{Experiment Design}
In our experimental design, we treat each dataset as a target domain in turn to assess UDA for binary and multi-class classification. For each target domain considered, we employ single-source, combined-source, and multi-source training schemes. In binary melanoma-nevus classification, we first train classifiers using both single and combined sources for each target domain without deploying UDA, aiming to compare the performance of each single source with an aggregated source. Next, we implement UDA for both single-source and combined-source cases, utilizing the adversarial-based DANN and ADDA methods, which have been proven effective in prior studies for binary skin lesion classification \cite{fogelberg2023domain, CHAMARTHI2024101430}. Lastly, we explore the efficacy of multi-source UDA in binary classification by applying two representative methods, M$^3$SDA and MDAN, that are designed for multi-domain adaptation. To address class imbalance, a weighted random sampler is incorporated into both combined-source and multi-source training, assuming a uniform target label distribution, same as \cite{CHAMARTHI2024101430}. This technique ensures balanced class ratios within each training mini-batch. Performance is evaluated using the AUROC metric for binary classification.

For multi-class classification, we apply the same training strategies but exclude single-source UDA due to the possible absence of target classes from the source domain. Therefore, we focus on combined-source and multi-source UDA training, whose merits shine when single-source training underperforms due to label space mismatch. For this task, we use accuracy as the evaluation metric. This discrepancy in metrics is due to the impracticality of computing AUROC when classes are missing from either the source or target domain (i.e.,  label space mismatch).

Fairness assessment is conducted on three datasets--Fitzpatrick17k, ISIC2020, and PAD-UFES-20--for binary classification. The training and test splits from these datasets are combined into one dataset for evaluation to ensure sufficient representation of each group. In alignment with \cite{daneshjou2022disparities, groh2021evaluating}, we consider skin color in Fitzpatrick17k as a sensitive attribute, categorizing the data into three groups according to their Fitzpatrick Skin Type (FST): light skin tone (FST 1-2), medium skin tone (FST 3-4), and dark skin tone (FST 5-6). For ISIC2020 and PAD-UFES-20, age is regarded as a sensitive attribute and can be used to form two groups ($\le$ 30 and $>$30). The distribution of these sensitive attributes is detailed in Appendix \cref{sec:fairness}. We evaluate fairness using three common metrics (PQD, DPM, and EOM) in our analysis across five training strategies: single-source and combined-source training, both with and without UDA, in addition to multi-source UDA. It's important to note that in the single-source scenarios, we use the average fairness results from training on the other 5 datasets independently as the results. 

\begin{figure*}[tbp]
    \centering
    \includegraphics[width=16cm]{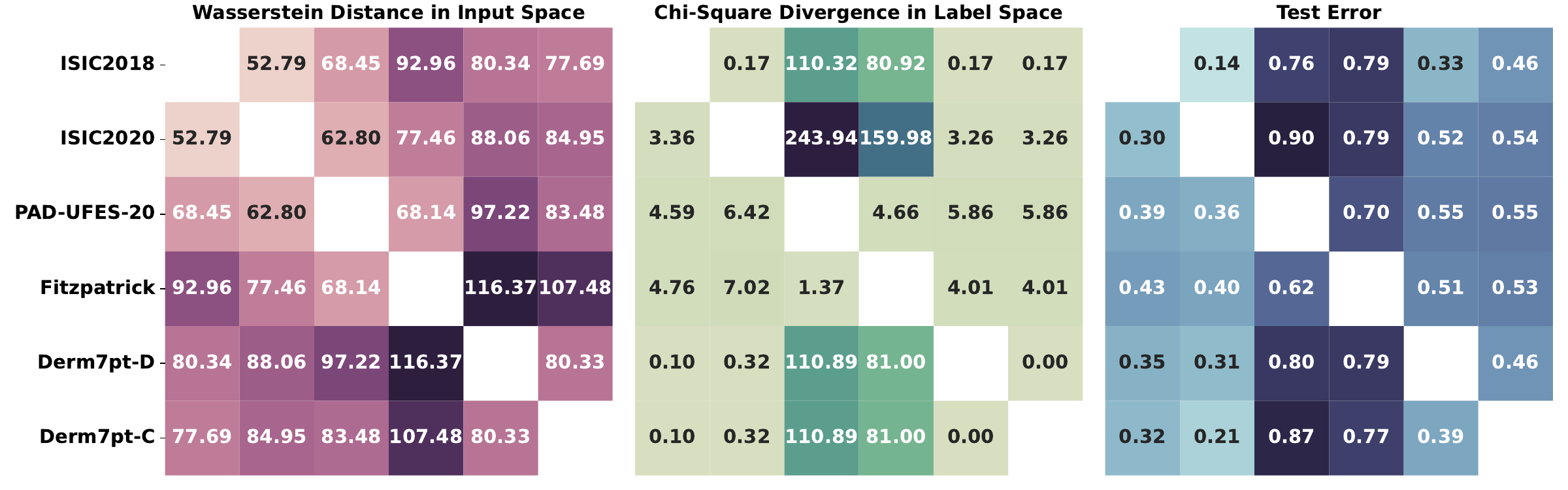}
    \caption{Feature distance (left), label distance (middle), and single-source test error for multi-class classification on the target test set (right). Pearson correlation coefficient is 0.31 between feature distance and test error and 0.78 between label distance and test error.}
    \label{fig:distances}
\end{figure*}

\begin{figure*}[!tbp]
    \centering
    \includegraphics[width=1.\textwidth]{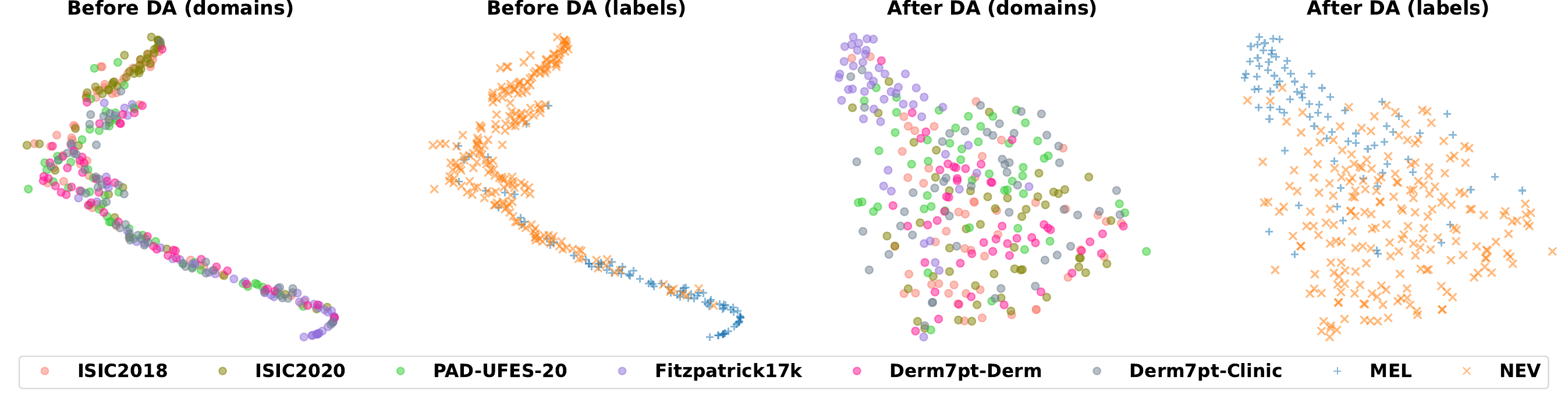}
    \caption{t-SNE figures of binary classification. Here target domain is fitz-roi and source domains are the other 5. Diagrams 1-2: combined-source without DA. Diagram 3-4: combined-source DANN. Diagrams are colored by domains (1, 3) and by labels (2, 4). After DANN, domains are well aligned and classes remain separable.}
    \label{fig:tsne_figure}
\end{figure*}

\subsection{Implementation Details}
All of our experiments utilized a pre-trained VGG16-BN\cite{simonyan2014very} in PyTorch as a feature extractor. Our methodology for single- and combined-source UDA experiments is based on a well-established repository\footnote{\href{https://github.com/thuml/Transfer-Learning-Library}{https://github.com/thuml/Transfer-Learning-Library}}. The default hyperparameters were employed for both single and combined sources, with the exception of the learning rates for single-source training, which were slightly decreased to $1 \times 10^{-4}$. As for the multi-source UDA, we utilized the official code with the default 
hyperparameters\footnote{\href{https://github.com/hanzhaoml/MDAN}{https://github.com/hanzhaoml/MDAN}} \footnote{\href{https://github.com/VisionLearningGroup/VisionLearningGroup.github.io/tree/master/M3SDA/code_MSDA_digit}{https://github.com/VisionLearningGroup/M3SDA}}. Each experiment was repeated for three times.
The data loading process was specifically adjusted to ensure the compatibility of these codes with our datasets. We deploy models on an RTX A6000 and an RTX 3090.
\section{Results}\label{sec:results}

\begin{figure*}[!tbp]
    \centering
    \includegraphics[width=1.\textwidth]{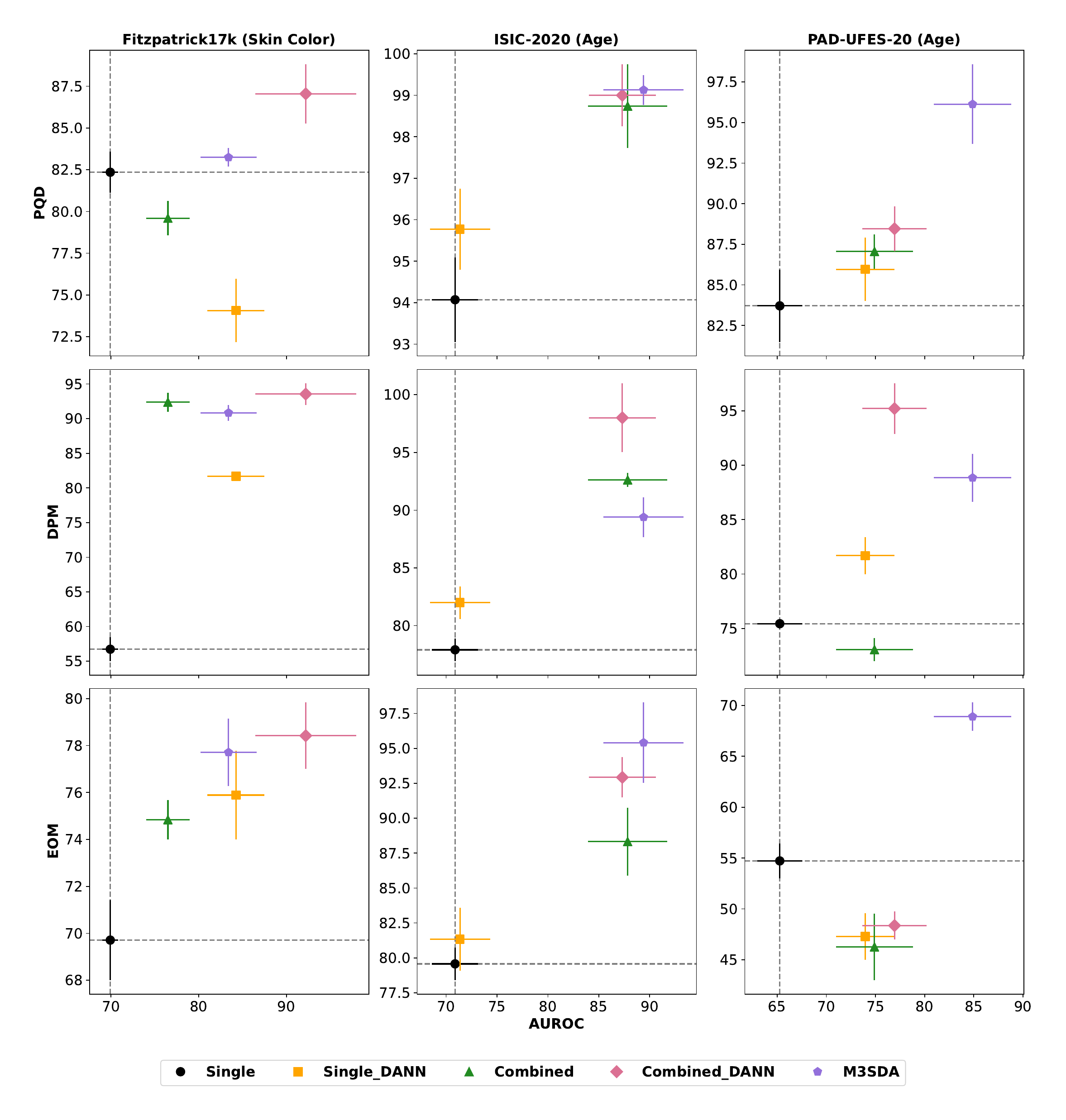}
    \caption{Fairness results on the Fitzpatrick17k, ISIC2020, and PAD-UFES-20 datasets for binary classification. Skin type is considered a sensitive attribute for Fitzpatrick17k, while age for ISIC2020 and PAD-UFES-20. All UDA-based methods in this experiment incorporate a weighted random sampler. Results can be found in Appendix \cref{fairness_table}}
    \label{fig:fairness_figure}
\end{figure*}

\subsection{Single Source vs. Multiple Sources}
Table \ref{binary_auroc_table} presents the results for binary classification, with each column representing results for a dataset as the target domain. An immediate observation is that leveraging multiple datasets yields the best results across all target domains. As highlighted in the table, the best results are either derived from combined-source or multi-source training schemes, even when data from the target domain isn’t available at all, validating the effectiveness of using multiple public datasets. The most prominent improvements are observed in cases where non-ISIC competition datasets are the target domain. Both ISIC2018 and ISIC2020 contain dermoscopic images and share similar class distribution that is highly skewed towards nevus (NEV) in binary classification. It is not surprising to observe that when each of them serves as the other’s source domain in single-source training, the results are sufficiently good due to their well aligned feature and label spaces. In contrast, the other four non-ISIC competition datasets benefit more from multiple sources, since they have distinct data distributions that can not be accommodated by a single dataset for training a reliable classifier. Considering single-source training alone, the classifier trained on the ISIC2018 dataset performs best on average across target domains, whereas using the PAD-UFES-20 dataset as a source yields the lowest average results. The performance gap may be attributed to the differences in dataset information: ISIC2018 covers 6 of 8 classes with the most examples, whereas PAD-UFES-20 has the fewest example. A similar pattern emerges in the more challenging multi-class task, as shown in Table \ref{8class_accuracy_table}. Since ISIC2018 and ISIC2020 share a similar distribution that skewed towards nevus (NEV) and melanoma (MEL), they can sufficiently serve as each other's training source and gain less advantage from multiple sources compared to other datasets.

To better understand these results, we use Wasserstein Distance~\cite{peyre2019computational, mehra2023analysis, tan2021otce} to measure the pixel-level difference and Chi-square divergence to measure label shift, for each single source and target pair (Fig.\ref{fig:distances}), in multi-class classification. Calculating their Pearson Correlations with the test error, we find that label shift is highly correlated with test error, yielding a Pearson Correlation coefficient as high as 0.78. The significant role of label shift is intuitive and anticipated. For instance, a classifier trained on ISIC 2020 performs uniformly poorly on other target domains, with the exception of ISIC 2018, which has a small feature and label distance from ISIC 2020. This result further confirms that the utilization of multiple sources can effectively mitigate label mismatch, especially when the distribution of the target domain is unknown.

\subsection{With vs. Without Domain Adaptation}
In binary classification, the single-source UDA method consistently outperforms its non-DA counterpart across all target domains in average AUROC. The only exception is when ISIC2018 serves as the target domain, where we observe a slight drop in average performance (see the last rows of the single and single-DANN sections in Table \ref{binary_auroc_table}). Performance improves further when UDA is applied to multiple sources, as highlighted in the previous section. Upon examining the results for combined-source and multi-source training more closely, we find that both strategies do not significantly outperform the ERM baseline (non-DA) on ISIC2018 and ISIC2020. As previously analyzed, this outcome is intuitive because these two ISIC datasets are each other’s sufficient training sets due to their similar feature and label distributions, with or without domain adaptation. Thus, including other smaller, non-ISIC datasets in the training set adds less informative value to the dominant ISIC source. In contrast, applying UDA on multiple sources consistently brings sizable improvements over the ERM baselines when non-ISIC datasets are the target domains. This indicates that useful information from public datasets has been effectively transferred to the target domain as a result of UDA. Taking Fitzpatrick17k as an example target domain, Fig.~\ref{fig:tsne_figure} demonstrates that applying DANN on a combined source forces the domain representations to integrate while preserving the separability of classes. For eight-class classification, despite the significantly increased task difficulty compared to binary classification, training on multiple sources with UDA generally outperforms non-DA training (Table \ref{8class_accuracy_table}).

\subsection{Impact on Fairness}
Fairness is tested on three datasets - Fitzpatrick17k, ISIC 2020, and PAD-UFES-20 - for the binary classification task, using three metrics: PQD, DPM, and EOM. According to Table \ref{binary_auroc_table}, DANN and M$^3$SDA with imbalance technique deliver the best average results in combined-source and multi-source training, respectively. Therefore, these two methods are selected to examine the impact of combined-source and multi-source training on fairness. As illustrated in Fig. \ref{fig:fairness_figure}, applying UDA on multiple sources consistently achieves superior classification performance while maintaining a minimal performance gap between groups. Despite the absence of fairness-focused techniques specifically aimed at minority groups, the results indicate that UDA effectively adapts lesion-related features and translates rich demographic information to the target domain, nudging the model towards a more equitable diagnostic system.
\section{Conclusion}\label{sec:conclusion}
In this study, we have explored three UDA training strategies: single-source, combined-source, and multi-source, when labeled target data is unavailable. The results validate the effectiveness of using multiple sources over a single source and establish the superiority of UDA over non-DA training in binary and multi-class classifications. Our results show that leveraging UDA with multiple sources enriches the class information of the training set and ensures integration of diverse domains. The merits of applying UDA on multiple sources particularly stand out when not a single dataset can cover the feature or label spaces of the target domain. Additionally, this study highlights the potential of utilizing UDA with multiple sources to promote fairness and mitigate bias against minority groups. We believe the practicality of our work is evident in the medical field, where custom datasets are usually too small to train a reliable diagnosis tool, and public datasets present a new, cost-free source of information. This work also opens up other interesting questions, such as the root cause of the improved fairness after adopting UDA on multiple sources, which will be left as a new direction for future work.
\section*{Acknowledgement}
This work was partly supported by the NSF EPSCoR-Louisiana Materials Design Alliance (LAMDA) program \#OIA-1946231 and partly by the Harold L. and Heather E. Jurist Center of Excellence for Artificial Intelligence at Tulane University.

{
    \small
    \bibliographystyle{ieeenat_fullname}
    \bibliography{main}
}

\clearpage
\setcounter{page}{1}
\maketitlesupplementary

\begin{figure}[htbp]
    \centering
    \includegraphics[width=1\linewidth, height=0.75\textheight, keepaspectratio]{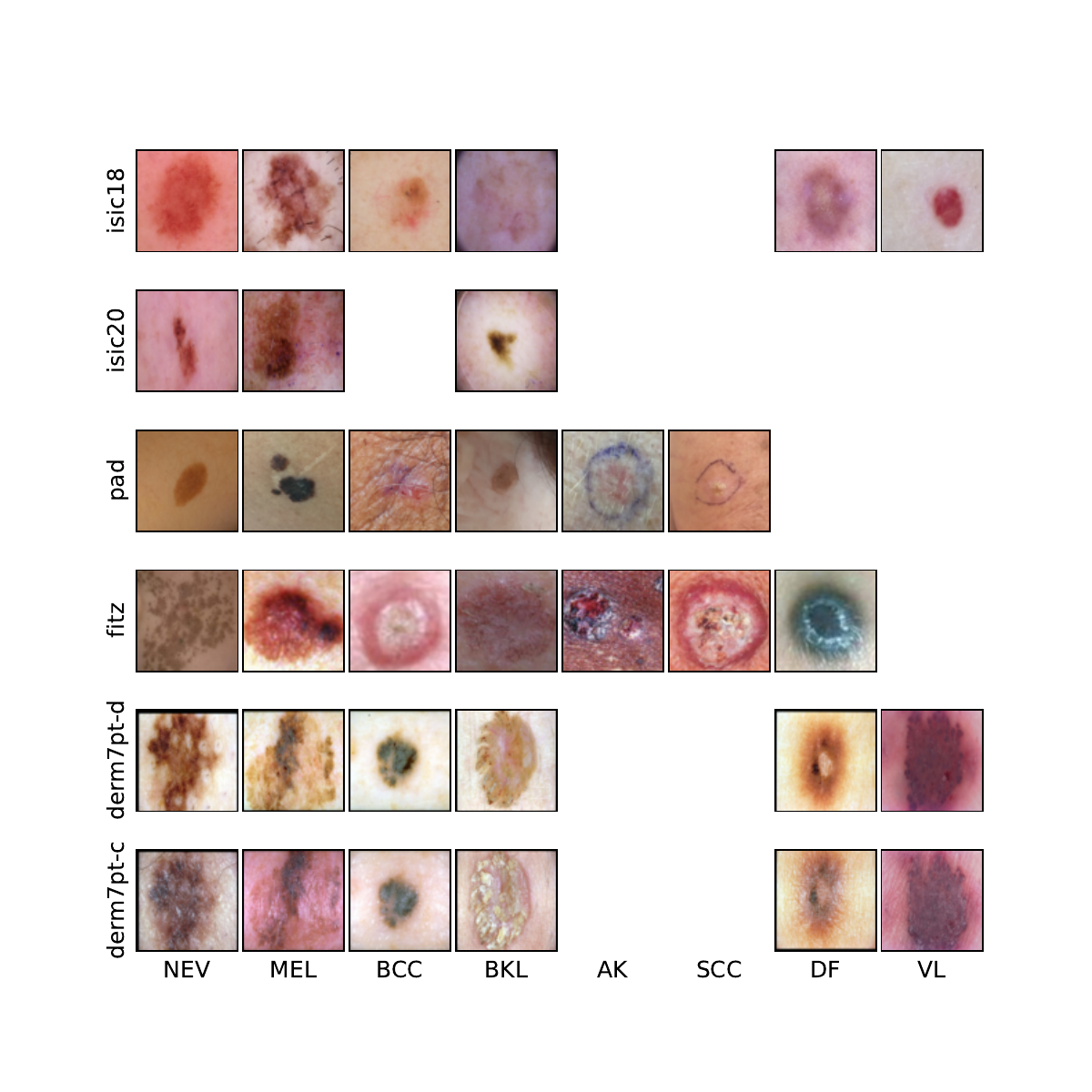}
    \caption{Image examples of each dataset considered in this study.}
    \label{fig:demo_imgs}
\end{figure}

\begin{figure}[htbp]
    \centering
    \flushleft
    \includegraphics[width=1\linewidth]{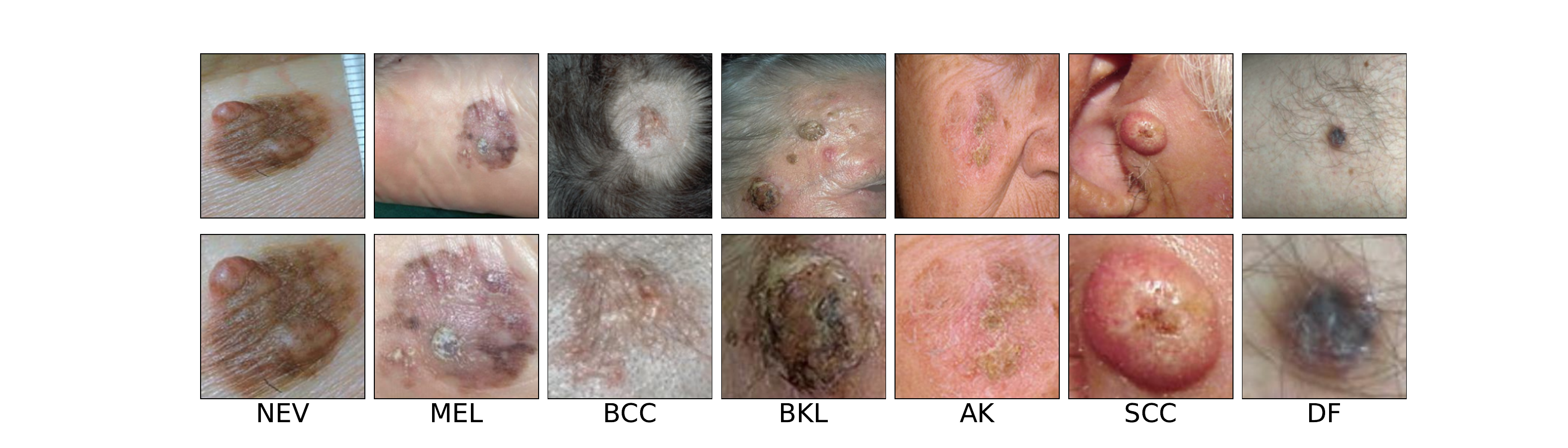}
    \caption{Examples of ROI-cropped vs. uncropped images for Fitzpatrick17k.}
    \label{fig:roi_demo}
\end{figure}

\section{Image Examples of Selected Datasets}
\label{sec:demo_imgs}
Fig. \ref{fig:demo_imgs} shows example images for each class and dataset considered in this study. Images from the Fitzpatrick17k dataset have been processed with ROI-based cropping. It is clear that images from different datasets are visually distinct, even for the same conditions, leading to a significant domain gap and posing challenges for classification.

\section{ROI Pre-processing}
\label{sec:roi}
In \cite{pmlr-v116-ghorbani20a}, region of interest (ROI) detection is utilized to separate skin lesions from clinical photos, effectively reducing noise and enhancing the lesion information ratio. In the context of our problem setting, dermoscopic images of skin lesions are mostly close-up shots centered on the lesions, whereas clinical photos are taken at varying distances from the lesions or from different angles. To minimize background noise, as well as the discrepancy between dermoscopic and clinical images, we fine-tune a YOLO-8 model, one of the SOTA ROI detection algorithms \cite{Jocher_YOLO_by_Ultralytics_2023}. After cropping, each image is resized to a resolution of $64\times 64$ pixels, which is the designated image size for subsequent experiments. Fig. \ref{fig:roi_demo} are examples of ROI-cropped images.

\section{Fairness Evaluation}
\label{sec:fairness}
\begin{figure}[htbp]
    \centering
    \includegraphics[width=1\linewidth]{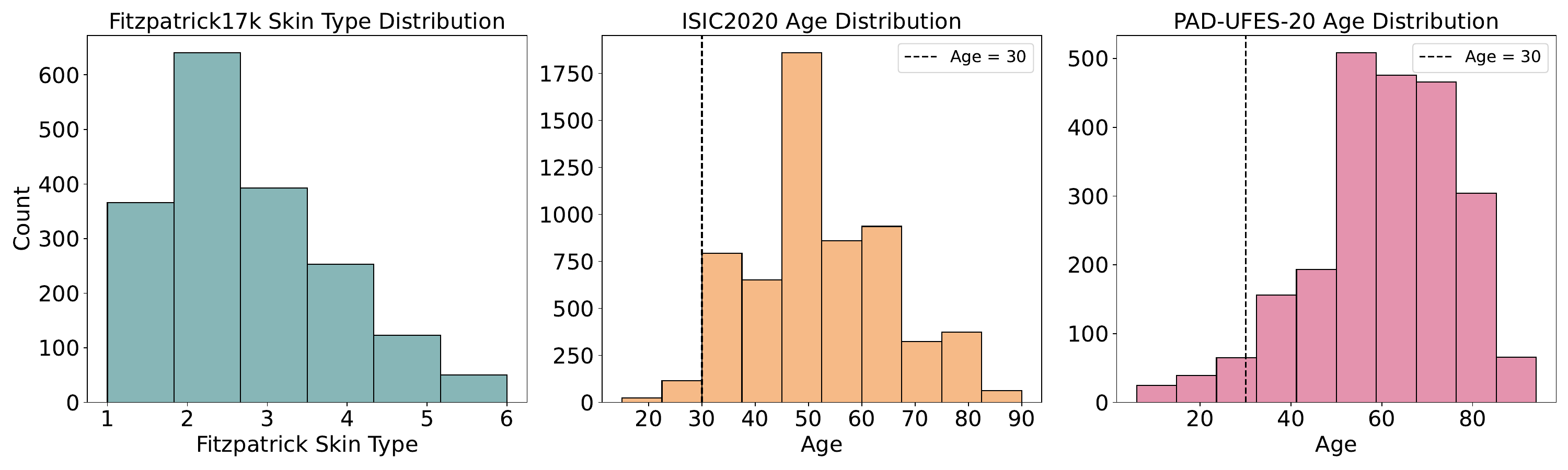}
    \caption{Sensitive Attribution Distribution for Datasets in Fairness Evaluation.}
    \label{fig:sensitive_attribute}
\end{figure}

Fig. \ref{fig:sensitive_attribute} shows the distribution of sensitive attributes across each dataset considered in the fairness evaluation. In the Fitzpatrick17k dataset, skin color is identified as a sensitive attribute. Notably, the dataset is skewed towards the light-skinned sub-population (FST 1-2), making images with dark skin tone (FST 5-6) a minority group. This imbalance may lead to under-diagnosis in dark-skinned individuals when using AI for early screening. Similarly, for ISIC2020 and PAD-UFES-20, age is considered a sensitive attribute, dividing the data into two groups. These datasets tend to be skewed towards individuals older than 30, potentially resulting in a higher risk of under-diagnosis in the younger population during AI-based early screening.

\begin{table}[!tbp]
    \centering
    \resizebox{0.5\textwidth}{!}{%
    \begin{tabular}{lcccccc}
    \toprule
    Domain & Metric & Single & Single DANN & Combined & Combined DANN & M$^3$SDA \\
    \midrule
    
    \multirow{3}{*}{fitz (skin color)} 
    & PQD & 82.4$\pm$\tiny{1.2} & 74.1$\pm$\tiny{1.9} & 79.6$\pm$\tiny{1.0} & \textbf{87.1$\pm$\tiny{1.8}} & 83.2$\pm$\tiny{0.6} \\
    & DPM & 56.7$\pm$\tiny{1.7} & 81.7$\pm$\tiny{0.7} & 92.4$\pm$\tiny{1.4} & \textbf{93.5$\pm$\tiny{1.6}} & 90.8$\pm$\tiny{1.1} \\
    & EOM & 69.7$\pm$\tiny{1.7} & 75.9$\pm$\tiny{1.9} & 74.8$\pm$\tiny{0.8} & \textbf{78.4$\pm$\tiny{1.4}} & 77.7$\pm$\tiny{1.4} \\
    \cmidrule{2-7}
    & AUROC & 69.9$\pm$\tiny{0.9} & 84.3$\pm$\tiny{3.3} & 84.0 $\pm$\tiny{2.5} & \textbf{92.2$\pm$\tiny{5.7}} & 83.4$\pm$\tiny{3.2} \\
    \midrule
    
    \multirow{3}{*}{isic2020 (age)} 
    & PQD & 94.1$\pm$\tiny{1.0} & 95.8$\pm$\tiny{1.0} & 98.7$\pm$\tiny{1.0} & 99.0$\pm$\tiny{0.8} & \textbf{99.1$\pm$\tiny{0.4}} \\
    & DPM & 77.9$\pm$\tiny{1.0} & 82.0$\pm$\tiny{1.4} & 92.6$\pm$\tiny{0.6} & \textbf{98.0$\pm$\tiny{3.0}} & 89.4$\pm$\tiny{1.7} \\
    & EOM & 79.6$\pm$\tiny{1.2} & 81.3$\pm$\tiny{2.3} & 88.3$\pm$\tiny{2.4} & 92.9$\pm$\tiny{1.4} & \textbf{95.4$\pm$\tiny{2.9}} \\
    \cmidrule{2-7}
    & AUROC & 70.9$\pm$\tiny{2.3} & 71.4$\pm$\tiny{3.0} & 87.8$\pm$\tiny{3.9} & 87.3$\pm$\tiny{3.3} & \textbf{89.4$\pm$\tiny{3.9}} \\
    \midrule
    
    \multirow{3}{*}{pad (age)} 
    & PQD & 83.7$\pm$\tiny{2.2} & 86.0$\pm$\tiny{2.0} & 87.1$\pm$\tiny{1.1} & 88.5$\pm$\tiny{1.4} & \textbf{96.1$\pm$\tiny{2.5}} \\
    & DPM & 75.4$\pm$\tiny{0.4} & 81.7$\pm$\tiny{1.7} & 73.1$\pm$\tiny{1.1} & \textbf{95.2$\pm$\tiny{2.3}} & 88.8$\pm$\tiny{2.2} \\
    & EOM & 54.7$\pm$\tiny{1.7} & 47.3$\pm$\tiny{2.3} & 46.3$\pm$\tiny{3.3} & 48.4$\pm$\tiny{1.4} & \textbf{68.9$\pm$\tiny{1.4}} \\
    \cmidrule{2-7}
    & AUROC & 65.3$\pm$\tiny{2.3} & 74.0$\pm$\tiny{3.0} & 74.9$\pm$\tiny{3.9} & 76.9$\pm$\tiny{3.3} & \textbf{84.9$\pm$\tiny{3.9}} \\
    \bottomrule
    \end{tabular}
    }
    \caption{Fairness Evaluation Metrics across Different Domains}
    \label{fairness_table}
\end{table}

\end{document}